\documentclass{article}

\usepackage{PRIMEarxiv}

\usepackage{cite}
\usepackage{amsmath,amssymb,amsfonts}
\usepackage{algorithm}
\usepackage{algorithmic}
\usepackage{graphicx}
\usepackage{textcomp}
\usepackage[table]{xcolor}
\usepackage[T1]{fontenc}
\usepackage{mathtools}

\usepackage{float}
\usepackage{placeins}
\usepackage{mathrsfs, bm}
\DeclareMathAlphabet{\mathcal}{OMS}{cmsy}{m}{n}
\DeclareSymbolFont{largesymbols}{OMX}{cmex}{m}{n}
\definecolor{gre}{RGB}{50, 168, 82}
\usepackage{tabularx}
\usepackage{booktabs}
\usepackage{multirow}
\usepackage{pifont}
\usepackage{array}
\usepackage{threeparttable}
\usepackage{bbding}
\usepackage{hyperref}
\def\BibTeX{{\rm B\kern-.05em{\sc i\kern-.025em b}\kern-.08em
    T\kern-.1667em\lower.7ex\hbox{E}\kern-.125emX}}
\newtheorem{proposition}{Proposition}

\title{Diffusion Plug-in Priors for Self-Supervised Ultrasound Video Restoration}

\author{
  \parbox{\textwidth}{\centering
    \setlength{\tabcolsep}{0pt}%
    \begin{tabular*}{0.9\textwidth}{@{\extracolsep{\fill}}ccccc}
      Chu Chen$^{1,*}$ & Mingduo He$^{2,*}$ & Yining Li$^{3,*}$ & Yinhao Wu$^{4}$ & Wei Tang$^{5,6}$
    \end{tabular*}
    \\[4pt]
    \begin{tabular*}{0.8\textwidth}{@{\extracolsep{\fill}}cccc}
      Yiwen Luo$^{7}$ & Yixuan Chen$^{5}$ & Pasquale Cascarano$^{8}$ & Kangning Cui$^{9,10,\dagger}$
    \end{tabular*}
    \\[10pt]
    \normalfont\small
    $^{1}$Department of Engineering, University of Cambridge, Cambridge, UK \\
    $^{2}$Department of Biomedical Engineering, School of Life Science and Technology, \\
    Huazhong University of Science and Technology, Wuhan, China \\
    $^{3}$Department of Intelligent Medical Engineering, School of Medicine, Nankai University, Tianjin, China \\
    $^{4}$University of Texas at Arlington, Arlington, TX, USA \\
    $^{5}$Department of Engineering Science, University of Oxford, Oxford, UK \\
    $^{6}$Oxford Suzhou Centre for Advanced Research, Suzhou, China \\
    $^{7}$Department of Electrical Engineering, City University of Hong Kong, Hong Kong SAR \\
    $^{8}$Department of the Arts, University of Bologna, Bologna, Italy \\
    $^{9}$Department of Data Science, City University of Hong Kong (Dongguan), Dongguan, China \\
    $^{10}$City University of Hong Kong, Hong Kong SAR \\[4pt]
    $^{*}$Equal contribution, listed in alphabetical order. \quad
    $^{\dagger}$Corresponding author: \texttt{kangning.cui@cityu-dg.edu.cn}
  }
}

\begin{document}
\maketitle

\begin{abstract}
Ultrasound videos suffer from hardware-constrained spatial resolution and signal-dependent speckle, while paired low- and high-resolution acquisitions are rarely available. We propose UltraDiff, a self-supervised framework for joint ultrasound video super-resolution and despeckling that uses a frozen pre-trained diffusion model as a plug-in prior. Rather than retraining the model, UltraDiff restores each frame by optimizing the sampler's latent seed under a range-constrained maximum a posteriori-inspired formulation. We derive a speckle-aware data-fidelity term from a Rayleigh scale-family model and characterize its relative-residual and locally variance-adaptive weighting properties. To address degeneracy near zero intensities, we introduce a stabilized surrogate fidelity that is bounded below and strictly convex in log intensity for positive observations in its unclamped form. Total-variation regularization complements the removal of mixed noise, and a temporal seed update warm-starts each frame from the optimized seed of its predecessor, accelerating the convergence. On EchoNet-LVH with simulated speckle and additive noise, UltraDiff with a natural-image prior improves peak signal-to-noise ratio by 1.24\,dB over the strongest baseline, and ultrasound-related priors yield further gains. On 100 real EchoNet-Dynamic videos, UltraDiff reduces the mean absolute error of zero-shot ejection fraction prediction from 7.31 to 6.37 percentage points compared with DUP. These results support ground-truth-free ultrasound video restoration using frozen diffusion priors and a speckle-aware fidelity, without paired training data or model retraining.
\end{abstract}

\keywords{ultrasound \and video super-resolution \and diffusion model \and self-supervised learning \and despeckling \and inverse problems}

\begin{figure}[t]
\centering
  \includegraphics[width=0.7\linewidth, trim=100pt 0pt 5cm 0pt, clip]{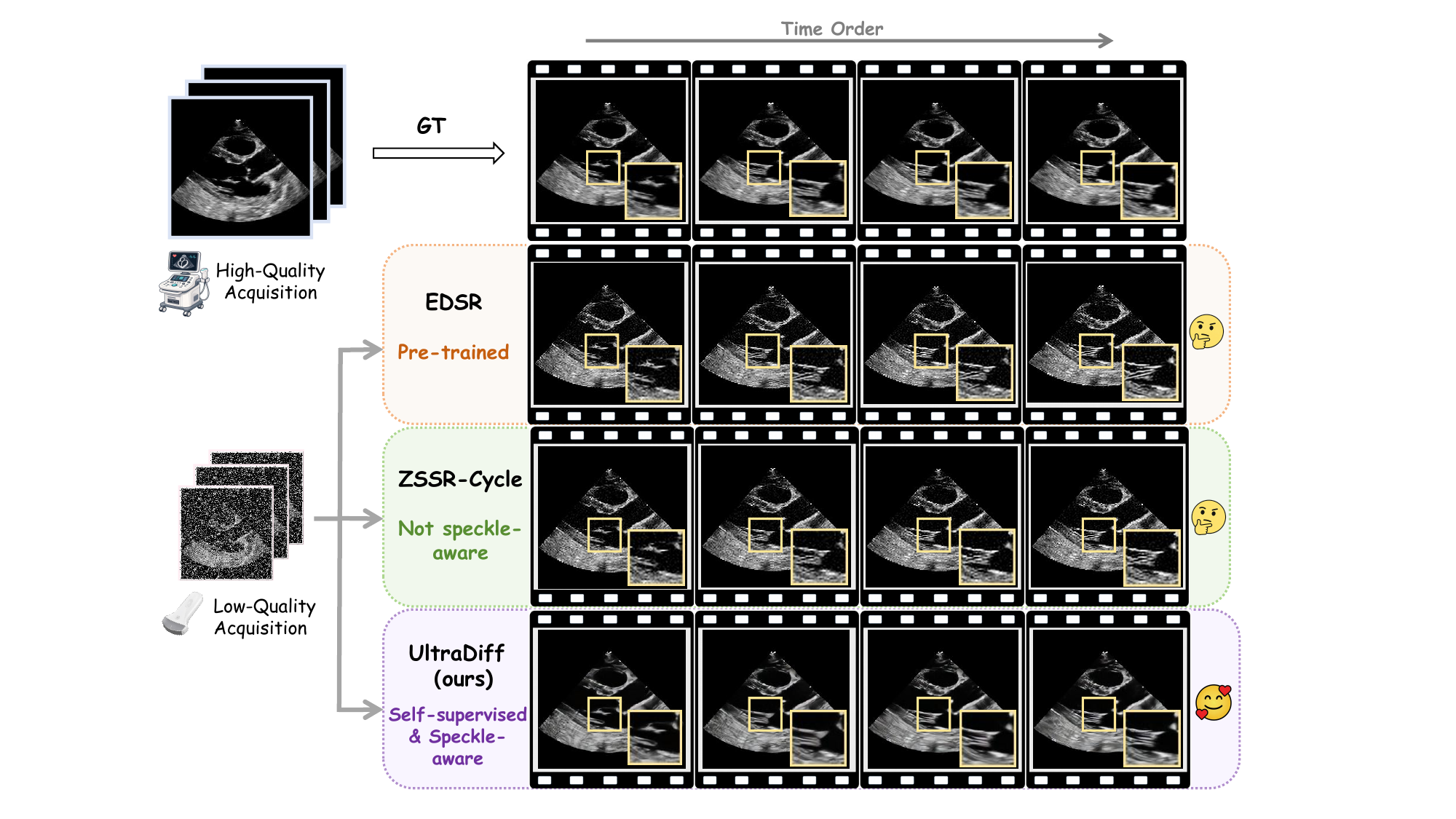}
  \caption{Motivating comparison of ultrasound video restoration. We compare reconstructed frames from LR ultrasound sequences with HR references and representative methods, including EDSR, ZSSR-Cycle, and UltraDiff. The zoom-in regions highlight structural details and tissue texture: UltraDiff suppresses noise while better preserving anatomical structures and the tissue echotexture.}
  \label{visualization}
\end{figure}

\section{Introduction}

Ultrasound (US) imaging is widely used in cardiology~\cite{aly2021cardiac,pellicori2021ultrasound,yang2025deep}, obstetrics~\cite{leung2021applications,alfirevic2015fetal}, and musculoskeletal imaging~\cite{patil2012role} because it is real-time, non-invasive, and cost-effective. Compared with static US images, US videos capture dynamic tissue motion, which is essential for applications such as breast lesion assessment~\cite{yoon2016validation,zhao2023deep} and cardiovascular monitoring~\cite{echolvh,christensen2024vision,yang2025latent}. In echocardiography, ejection fraction (EF) estimation depends on clear cardiac structures throughout the cardiac cycle.

Despite these advantages, the quality of US videos is inherently constrained by the physics of acoustic imaging. Spatial resolution is limited by the frequency-penetration trade-off of transducer design, while image quality is further affected by signal-dependent speckle noise caused by the interference of backscattered waves~\cite{sivaanpu2025speckle}. Classical despeckling methods reduce these fluctuations~\cite{michailovich2006despeckling,sivaanpu2025speckle}, but must balance noise suppression against the loss of tissue echotexture, which itself carries diagnostic information. These degradations reduce boundary visibility and can compromise downstream measurements such as automated EF prediction. Super-resolution (SR) offers a computational route beyond hardware-limited resolution~\cite{yue2016image}. However, robust US video restoration remains challenging: paired low-resolution (LR) and high-resolution (HR) US videos are rarely available, acquisition conditions vary across devices and operators, and US noise statistics differ substantially from natural-image degradations. As illustrated in Fig.~\ref{visualization}, representative SR baselines either over-smooth cardiac structures or fail to recover realistic tissue texture.

\subsection{Related Work and Motivation}

SR methods for natural images and videos have advanced rapidly with large-scale paired datasets~\cite{agustsson2017ntire,nah2019ntire,liu2013bayesian}. Single-image SR methods commonly use deep convolutional networks~\cite{lim2017enhanced,zhang2018residual}, while video SR methods incorporate temporal information through flow-based~\cite{xue2019video} or attention-based mechanisms~\cite{zhou2024video}. However, these models are typically trained on fixed degradation distributions and may generalize poorly to clinical US videos, where image appearance, resolution, and speckle patterns vary across acquisition settings.

In US image SR, paired LR and HR data of the same anatomy are particularly difficult to obtain, so existing studies rely on simulated pairs or self-supervised strategies. Choi et al.~\cite{Choi2018IUS} adapted SRGAN~\cite{Ledig_2017_CVPR} for B-mode US image enhancement, while a U-Net-style architecture~\cite{van2019deep} was trained on simulated pairs for super-resolved US reconstruction. Liu et al.~\cite{liu2021perception} proposed ZSSR-Cycle, a zero-shot and self-supervised generative adversarial framework for US image SR. A related line of work exploits the implicit prior of network architectures: Deep Image Prior (DIP)~\cite{Ulyanov_2018_CVPR} fits a randomly initialized network to a single degraded image, and its video extensions propagate network weights or priors across frames~\cite{lei2022deep,cascarano2021recursive}. Following this paradigm, DUP~\cite{chen2025blind} performs video-adaptive, ground-truth-free US video quality improvement. These methods re-fit network parameters for every input and rely on early stopping to avoid fitting noise. Moreover, they do not explicitly couple a learned generative prior with a speckle-aware US measurement model.

Diffusion models (DMs) have been widely used as generative priors for image restoration, improving perceptual detail through iterative denoising~\cite{song2020denoising,ho2020denoising}. Diffusion-based methods have been explored for video enhancement~\cite{li2025diffvsr} and US image SR~\cite{liu2024super}. DMPlug~\cite{wang_dmplug_2024} shows that a pre-trained DM can be used as a plug-in prior for inverse problems without task-specific retraining, by optimizing the input of the sampler under a least-squares fidelity on single images. Applying such priors to US video restoration, however, is non-trivial: (i) a least-squares fidelity implicitly assumes signal-independent noise, which is inconsistent with multiplicative speckle; (ii) the method must adapt to each LR video without paired HR supervision; and (iii) temporal redundancy should be exploited without training additional temporal modules.

\subsection{Contributions}

To address these challenges, we propose UltraDiff, a self-supervised framework for ultrasound video super-resolution and despeckling. UltraDiff uses a frozen pre-trained diffusion model as a plug-in prior and adapts it to each LR video by optimizing latent seeds rather than model parameters. The optimized HR frames are constrained by a degradation operator and a speckle-aware data-fidelity term, so restoration is guided by both the learned image prior and the ultrasound measurement model. The main contributions of this work are summarized as follows:

\begin{itemize}
    \item \textit{We propose UltraDiff}, a self-supervised US video restoration framework for joint super-resolution and despeckling that requires neither paired HR videos nor DM retraining.
    \item \textit{We develop a speckle-aware restoration objective} based on a Rayleigh scale-family model and a range-constrained MAP-inspired formulation. We characterize its relative-residual behavior and local variance-adaptive weighting, and introduce a stabilized fidelity with favorable optimization properties.
    \item \textit{We introduce a temporal seed update (SUP) strategy} that warm-starts each frame from the optimized seed of its predecessor, combining the previous optimized seed with Gaussian perturbations and reducing the iteration budget for subsequent frames.
    \item \textit{We validate UltraDiff on two public echocardiography datasets}, showing improved restoration quality, noise robustness, and zero-shot EF prediction over representative SR baselines, with ablations verifying key components.
\end{itemize}

\section{Methodology}

\begin{figure*}[t]
\centering
  \includegraphics[width=\linewidth, trim=20pt 100pt 0.5cm 75pt]{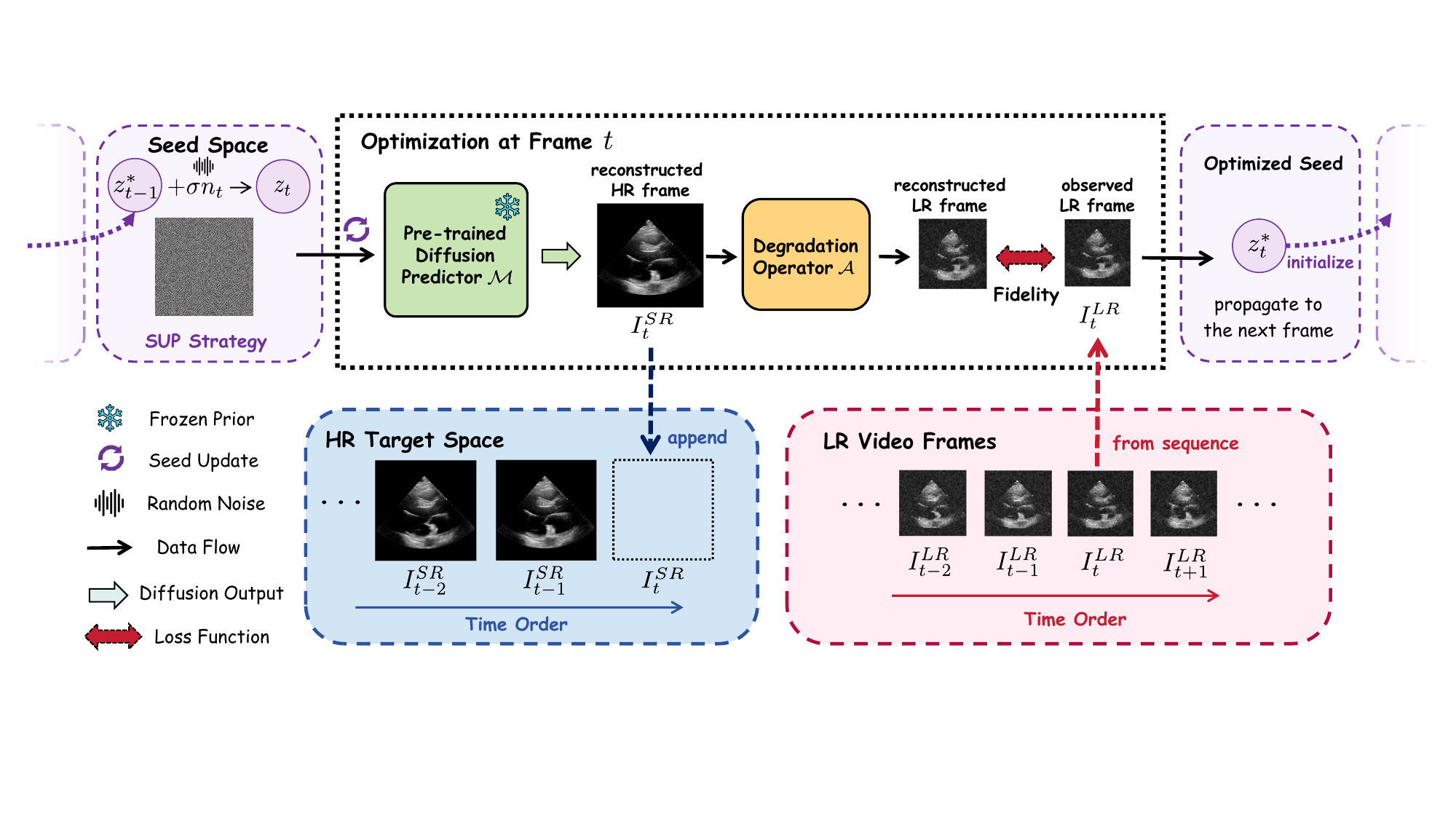}
  \caption{Overview of the UltraDiff framework. A frozen pre-trained diffusion model maps latent seeds to HR estimates, which are degraded back to the LR space and optimized against the observed LR frames. The optimized seeds are propagated and perturbed across consecutive frames to enable video-adaptive restoration without diffusion model retraining.}
  \label{flowchart}
\end{figure*}

\subsection{Problem Formulation}
\label{sec:model}
We consider frame-wise restoration of an LR US video $\{I_t^{LR}\}_{t=1}^{T}$ under the observation model~\cite{sivaanpu2025speckle}
\begin{equation}
    I_t^{LR} = (\mathcal{A} I_t^{HR}) \odot \phi_t + \psi_t, \quad t=1,2,\dots,T,
    \label{prob}
\end{equation}
where $\odot$ denotes element-wise multiplication, $\mathcal{A} \in \mathbb{R}^{MN \times s^2MN}$ is a linear downsampling operator, $I_t^{LR} \in \mathbb{R}^{MN}$ and $I_t^{HR} \in \mathbb{R}^{s^2MN}$ are the vectorized $t$-th LR and HR frames with intensities normalized to $[0,1]$, $s$ is the upscaling factor, and $T$ is the number of frames. The terms $\phi_t, \psi_t \in \mathbb{R}^{MN}$ denote multiplicative speckle and additive noise, respectively, with i.i.d.\ entries, mutually independent and independent of $I_t^{HR}$, $\mathbb{E}[\phi_t(i)]=1$, and $\mathbb{E}[\psi_t(i)]=0$. For brevity, we write $y_i \coloneqq I_t^{LR}(i)$ and $x_i \coloneqq [\mathcal{A} I_t^{HR}]_i$ and drop the frame index when no ambiguity arises. Under~\eqref{prob}, $\mathbb{E}[y_i \mid x_i]=x_i$ and
\begin{equation}
    \mathrm{Var}[y_i \mid x_i] = \sigma_\phi^2\, x_i^2 + \sigma_\psi^2,
    \label{eq:var}
\end{equation}
where $\sigma_\phi^2$ and $\sigma_\psi^2$ are the variances of $\phi_t(i)$ and $\psi_t(i)$. The first term is signal-dependent, while the second term dominates where $x_i$ is small, e.g., in anechoic cavities and outside the imaging sector.

We use~\eqref{prob} as a practical observation model for normalized B-mode video frames, which combines signal-dependent speckle with additive perturbations~\cite{michailovich2006despeckling,sivaanpu2025speckle}. The multiplicative component motivates a scale-family likelihood $p(y_i\mid x_i)=x_i^{-1}f(y_i/x_i)$, which provides the basis for our speckle-aware data fidelity. The diffusion generator and TV regularization further constrain the reconstruction under mixed-noise degradation (Section~\ref{sec:prior}).

\subsection{Speckle-Aware Data Fidelity}
\label{sec:fidelity}
To construct a tractable speckle-aware fidelity, we start from the Rayleigh model of fully developed envelope speckle, whose distribution scale is proportional to local reflectivity~\cite{sivaanpu2025speckle}:
\begin{equation}
    p(y_i \mid x_i) = \frac{y_i}{\beta^2 x_i^2} \exp\!\left( -\frac{y_i^2}{2\beta^2 x_i^2} \right), \quad y_i \geq 0,
    \label{eq:rayleigh}
\end{equation}
where $\beta>0$ links the Rayleigh scale to $x_i$; since $\mathbb{E}[y_i \mid x_i] = \beta\sqrt{\pi/2}\,x_i$, the choice $\beta=\sqrt{2/\pi}$ yields unit-mean speckle. Eq.~\eqref{eq:rayleigh} has the scale-family form with $f(r) = (r/\beta^2)\exp(-r^2/(2\beta^2))$. The negative log-likelihood of a frame is
\begin{equation}
-\log p(I_t^{LR}\mid I_t^{HR})
= \sum_{i=1}^{MN} \ell(x_i; y_i) + \sum_{i=1}^{MN} \bigl(2\log\beta - \log y_i\bigr),
\label{eq:neg_log_likelihood}
\end{equation}
where the second sum does not depend on $I_t^{HR}$ and the per-pixel term is $\ell(x;y) \coloneqq 2\log x + y^2/(2\beta^2x^2)$. Its structure is characterized as follows (proofs are given in the Appendix).

\begin{proposition}
\label{prop:fidelity}
Let $y>0$ and $x>0$. \\
\textit{(i) Log-residual form:} with $r = \log y - \log x$,
\begin{equation}
    \ell(x;y) = 2\log y + \rho(r), \quad \rho(r) = \frac{e^{2r}}{2\beta^2} - 2r,
    \label{eq:logres}
\end{equation}
where $\rho$ is strictly convex. Hence $\ell$ has the unique minimizer $\bar{x} = y/(\sqrt{2}\beta)$ and is scale-equivariant, i.e., $\ell(cx;cy) = \ell(x;y) + 2\log c$ for all $c>0$. \\
\textit{(ii) Local weighting:} around $\bar{x}$,
\begin{equation}
    \ell(x;y) = \ell(\bar{x};y) + \frac{2}{\bar{x}^2}(x-\bar{x})^2 + O(|x-\bar{x}|^3).
    \label{eq:wls}
\end{equation}
\end{proposition}

Property (i) shows that $\ell$ penalizes the log-ratio between observation and prediction rather than their difference, so a given absolute error costs less in a bright myocardial region than in a dark one, in agreement with the proportionally larger fluctuations of bright regions in~\eqref{eq:var}. Property (ii) makes this precise: near its minimizer, $\ell$ is a weighted least-squares term with weight $\propto 1/\bar{x}^2$, i.e., inversely proportional to the speckle variance $\sigma_\phi^2 x^2$. By contrast, the common mean squared error (MSE) (e.g., used in DMPlug~\cite{wang_dmplug_2024}) has constant curvature and implicitly assumes signal-independent noise, so it over-weights bright, high-variance regions and encourages fitting speckle there. In the log-intensity domain, $\ell$ is an additive penalty $\rho$ on the log-residual, i.e., the negative log-likelihood of log-Rayleigh noise. Proposition~\ref{prop:fidelity} characterizes the per-pixel geometry only. The overall objective remains nonconvex in the latent seed.

Two issues prevent using $\ell$ directly. First, zero-valued observations commonly occur in masked background regions and anechoic cavities. For such pixels, the reduced fidelity $\ell(x;0)=2\log x$ decreases without bound as $x\to 0$, causing excessive influence from background pixels during optimization. Second, the weight $2/\bar{x}^2$ in~\eqref{eq:wls} diverges in dark regions, where by~\eqref{eq:var} the additive term dominates, and the speckle likelihood is misspecified. We therefore use the stabilized fidelity
\begin{equation}
\mathcal{L}_{\text{data}}(I_t^{LR}, I)
= \frac{1}{MN}\sum_{i=1}^{MN} \ell_s\bigl([\mathcal{A}I]_i;\, y_i\bigr),
\label{eq:fidelity_final}
\end{equation}
with $\ell_s(x;y) = 2\log(1+x) + y^2/(2\beta^2 \max(x,\epsilon)^2)$, where the clamp at a small $\epsilon>0$ avoids division by zero.

\begin{proposition}
\label{prop:stab}
For $x>0$ and $y\geq 0$ (ignoring the clamp): \\
(i) $\ell_s(x;y)\geq 0$, and the gradient of its logarithmic term is bounded by $2$. \\
(ii) For $y>0$, $\ell_s$ is strictly convex in $u=\log x$ and has a unique stationary point $x^*$, the positive root of $2x^3 = \beta^{-2}y^2(1+x)$; for $\beta=1$ and $y\in(0,1]$, $y \leq x^* \leq 1$, with $x^*=y$ if and only if $y=1$. \\
(iii) At $x^*$, $\partial_x^2 \ell_s = (6+4x^*)/\bigl(x^*(1+x^*)^2\bigr)$, which decreases monotonically in $x^*$ and grows as $O(1/x^*)$ when $x^*\to 0$, compared with $4/\bar{x}^2$ for $\ell$ and a constant for the MSE.
\end{proposition}

Thus, $\ell_s$ removes the degeneracy of $\ell$ while retaining its variance-adaptive behavior: bright pixels are still down-weighted relative to dark ones, but the dark-region weight grows as $O(1/x)$ instead of $O(1/x^2)$, interpolating between multiplicative-noise and homoscedastic weighting in line with the mixed structure of~\eqref{eq:var}. For $\beta=1$ and $y\in(0,1]$, the per-pixel stationary point lies in $[y,1]$, reflecting the intensity-dependent shift introduced by stabilization. The estimate, however, is not determined pixel-wise: $\mathcal{A}$ couples neighboring HR pixels, the HR frame is restricted to the generator (Section~\ref{sec:prior}), and the optimization is early-stopped, so the reconstruction is governed jointly by fidelity, prior, and regularizer (accuracy is assessed in Section~\ref{sec:experiments}). In the stabilized fidelity, we set $\beta=1$ to fix the relative scaling between the logarithmic and signal-dependent terms throughout all experiments.

\subsection{Diffusion Plug-in Prior and Overall Objective}
\label{sec:prior}
Fig.~\ref{flowchart} summarizes the framework. Let $\mathcal{M}:\mathbb{R}^{d}\to\mathbb{R}^{s^2MN}$ denote the deterministic $K$-step DDIM sampler~\cite{song2020denoising} of a frozen pre-trained DM, which maps an initial latent seed $z$ to an image, followed by an affine rescaling to $[0,1]$. Following DMPlug~\cite{wang_dmplug_2024}, we treat $\mathcal{M}$ as a fixed generator and parameterize the HR frame as $I_t^{HR}=\mathcal{M}(z)$. Combining the stabilized data fidelity with a prior that restricts the HR frame to the range $\mathcal{R}(\mathcal{M}) = \{\mathcal{M}(z): z\in\mathbb{R}^d\}$ and penalizes its total variation (TV), we obtain the range-constrained MAP-inspired optimization problem
\begin{equation}
    \hat{I}_t = \operatorname*{argmin}_{I\in\mathcal{R}(\mathcal{M})} \mathcal{L}_{\text{data}}(I_t^{LR}, I) + \lambda R(I),
    \label{eq:map}
\end{equation}
which is solved in the latent space as
\begin{equation}
    z_t^* = \operatorname*{argmin}_{z}\, \mathcal{L}_{\text{data}}(I_t^{LR}, \mathcal{M}(z)) + \lambda R(\mathcal{M}(z)),
\label{eq:optim}
\end{equation}
with $\hat{I}_t = \mathcal{M}(z_t^*)$. Here, $\lambda$ controls the regularization strength and $R$ is the anisotropic TV~\cite{rudin1992nonlinear}
\begin{equation}
    R(I) = \frac{1}{s^2MN}\sum_{i=1}^{s^2MN}\bigl(|(\Delta_h I)_i| + |(\Delta_v I)_i|\bigr),
    \label{eq:tv}
\end{equation}
where $\Delta_h$ and $\Delta_v$ are first-order horizontal and vertical finite-difference operators. Both terms in~\eqref{eq:optim} are normalized per pixel, so $\lambda$ is comparable across resolutions. The DM weights remain fixed, and $\nabla_z$ is obtained by backpropagation through the $K$ sampler steps.

\subsection{Temporal Seed Update}
\label{sec:sup}
For a video, \eqref{eq:optim} is solved frame by frame. Rather than initializing each seed independently ($z_t = n_t$, $n_t \sim \mathcal{N}(0,I)$), which discards temporal context, or reusing a fixed seed, which limits frame-specific adaptation, the seed update (SUP) strategy initializes
\begin{equation}
z_t^{(0)} = 
\begin{cases}
    n_1, & t=1,\\
    z_{t-1}^* + \sigma n_t, & t \geq 2,
\end{cases}
\label{eq:sup}
\end{equation}
where $\sigma$ controls the perturbation magnitude. We offer three remarks on this design.

\textit{Perturbation strength.} With $\sigma=0$, the previous optimized seed is reused without fresh exploration, while as $\sigma$ grows, fresh noise contributes more strongly to the initialization. SUP with small $\sigma$ balances information reuse and frame-specific adaptation.

\textit{Bounded initial discrepancy.} Since $\mathcal{M}$ is a composition of $K$ evaluations of a smooth network, it is locally Lipschitz: with a local constant $L$ around $z_{t-1}^*$, $\|\mathcal{M}(z_t^{(0)}) - \hat{I}_{t-1}\|_2 \leq L\sigma\|n_t\|_2$. Although $L$ is unknown and possibly large, the initial discrepancy thus scales with $\sigma$, in line with the observation that nearby DDIM latents map to similar images~\cite{song2020denoising}. Because consecutive echocardiographic frames differ only slightly at typical frame rates~\cite{echodynamic}, this warm start motivates the smaller iteration budget for subsequent frames (Section~\ref{sec:implementation}).

\textit{Implicit temporal coupling.} For a locally quadratic objective, finite-step optimization retains a dependence on its initialization $z^{(0)}$. Hence, SUP with a finite budget and early stopping implicitly biases $z_t^*$ toward $z_{t-1}^*$, favoring temporally coherent reconstructions without an explicit temporal penalty or motion estimation, while the perturbation $\sigma n_t$ keeps the optimization from being locked in the previous frame-specific content, such as wall and valve motion, which can be fitted.

\subsection{Early Stopping and Algorithm}
\label{sec:es}
Iteratively fitting a highly expressive generator eventually reproduces measurement noise, as observed for DIP~\cite{Ulyanov_2018_CVPR}, so we stop each frame when the objective plateaus. Let $\mathcal{L}^{(k)}$ be the objective of~\eqref{eq:optim} at iteration $k$ and $\mathcal{L}_{\text{best}}$ the lowest value observed so far. After a warm-up of $K_{\min}$ iterations, a counter is incremented whenever $\mathcal{L}^{(k)} \geq (1-\epsilon_s)\mathcal{L}_{\text{best}}$ and reset otherwise. Optimization stops when the counter reaches the patience $P$ or after $K_{\max}$ iterations. Because SUP provides a warm start, subsequent frames use smaller $K_{\min}$ and $K_{\max}$ than the first frame. The complete procedure is summarized in Algorithm~\ref{alg:ultradiff}.

\begin{algorithm}[t]
\caption{UltraDiff for an LR ultrasound video}
\label{alg:ultradiff}
\begin{algorithmic}[1]
\REQUIRE $\{I_t^{LR}\}_{t=1}^T$, $\mathcal{M}$, $\mathcal{A}$, $\lambda$, $\sigma$, $\eta$, $\epsilon$, $(K_{\min}^{(t)}, K_{\max}^{(t)})$, $P$, $\epsilon_s$
\ENSURE HR frames $\{\hat{I}_t\}_{t=1}^T$
\FOR{$t=1$ \TO $T$}
\STATE Initialize $z$ by~\eqref{eq:sup}; $\mathcal{L}_{\text{best}}\leftarrow\infty$; $c\leftarrow 0$
\FOR{$k=1$ \TO $K_{\max}^{(t)}$}
\STATE $\mathcal{L} \leftarrow \mathcal{L}_{\text{data}}(I_t^{LR},\mathcal{M}(z)) + \lambda R(\mathcal{M}(z))$
\STATE $z \leftarrow \mathrm{Adam}(z, \nabla_z \mathcal{L}, \eta)$
\IF{$\mathcal{L} < (1-\epsilon_s)\mathcal{L}_{\text{best}}$}
\STATE $\mathcal{L}_{\text{best}}\leftarrow \mathcal{L}$; $c\leftarrow 0$
\ELSIF{$k > K_{\min}^{(t)}$}
\STATE $c \leftarrow c+1$; \textbf{if} $c \geq P$ \textbf{then break}
\ENDIF
\ENDFOR
\STATE $z_t^* \leftarrow z$; \ $\hat{I}_t \leftarrow \mathcal{M}(z_t^*)$
\ENDFOR
\end{algorithmic}
\end{algorithm}

\begin{figure*}[t]
\centering
  \includegraphics[width=\linewidth, trim=0pt 140pt 0cm 100pt, clip]{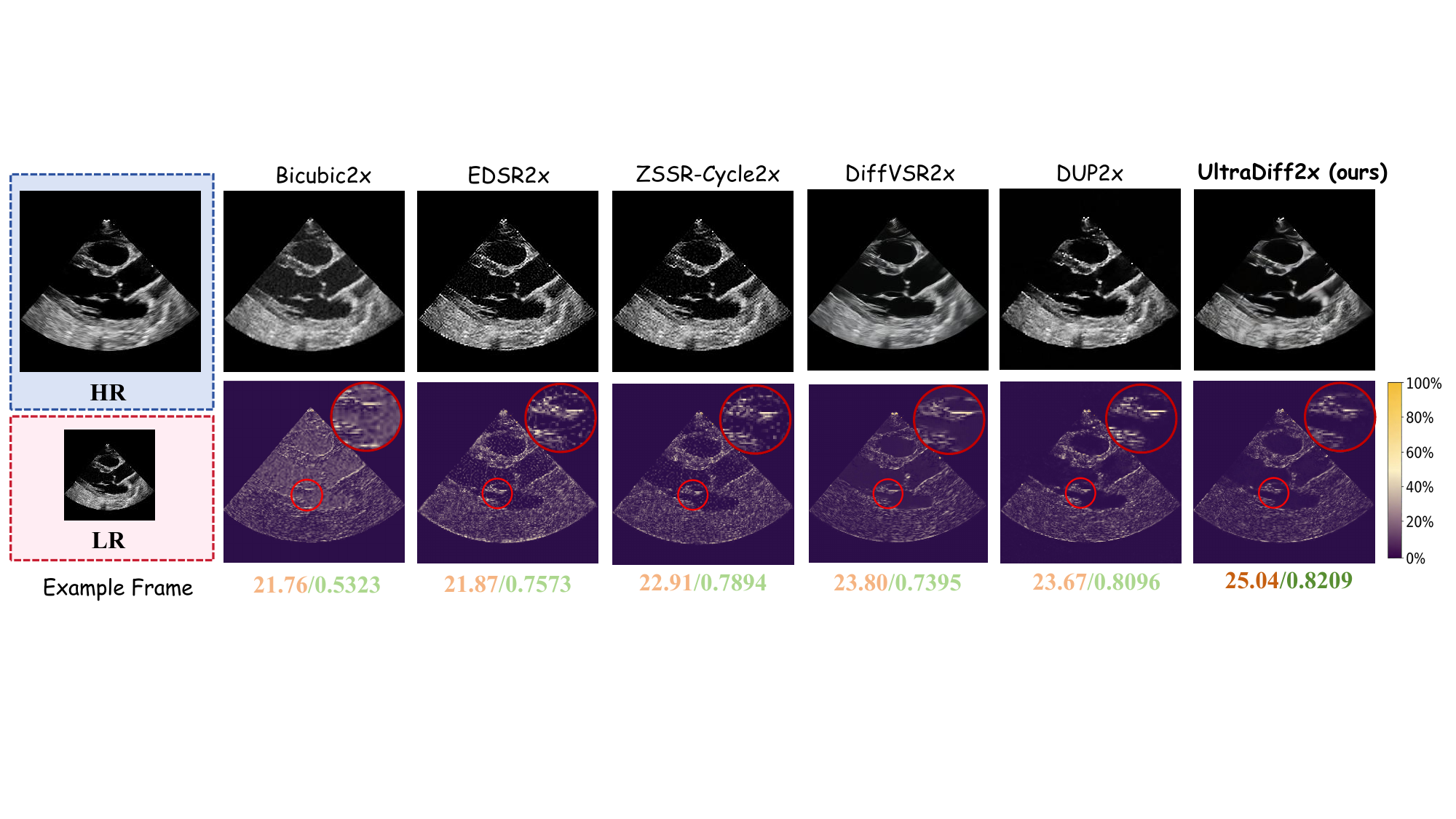}
  \caption{Qualitative comparison for 2$\times$ ultrasound video super-resolution. The left panel shows the HR reference and the simulated LR input. The top row compares SR outputs from different methods, with \textcolor[HTML]{C45E10}{PSNR}\textcolor[HTML]{578C31}{/SSIM} reported below each result, and the bottom row shows the corresponding error maps with zoomed regions for structural inspection.}
  \label{fig:sr_visual}
\end{figure*}

\section{Experiments}
\label{sec:experiments}
We evaluate UltraDiff in terms of restoration quality and robustness on simulated LR--HR echocardiography pairs, component contributions through ablations, and downstream utility through zero-shot EF prediction on real clinical videos.

\begin{figure*}[t]
    \centering
    \includegraphics[width=\linewidth, trim=15pt 65pt 0cm 0cm, clip]{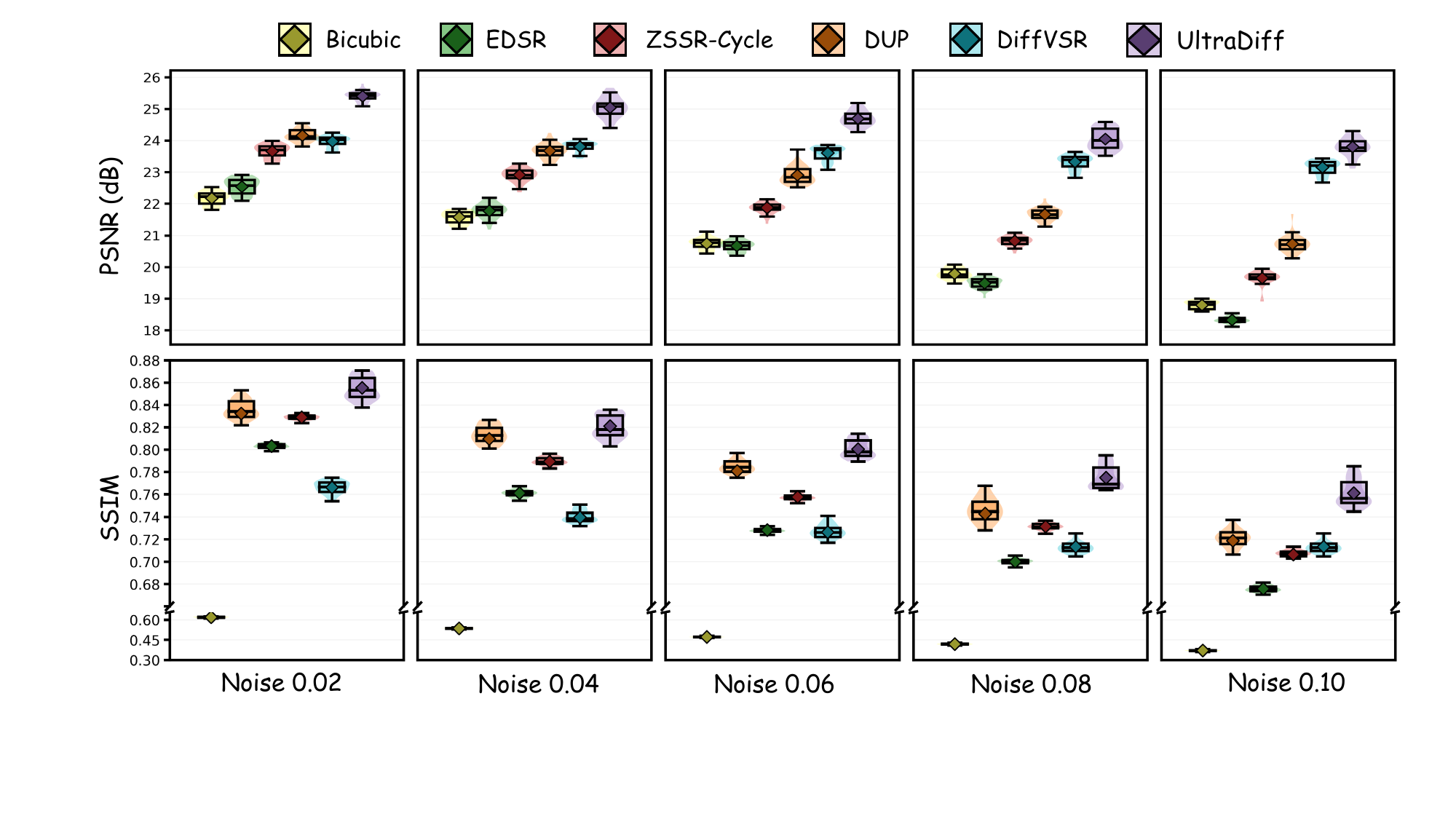}
    \caption{Robustness analysis under increasing mixed-noise levels from 0.02 to 0.10. Box-and-violin plots report PSNR and SSIM across different methods. UltraDiff maintains higher medians and more concentrated distributions, indicating stronger restoration stability under heavier degradation.}
    \label{fig:robust2noise}
\end{figure*}

\begin{table*}[t]
\centering
\caption{Quantitative restoration performance on EchoNet-LVH. UltraDiff variants use different pre-trained diffusion priors. The best, second-best, and third-best results are highlighted in \textcolor[HTML]{C00000}{red}, \textcolor[HTML]{E46C0A}{orange}, and \textcolor[HTML]{0070C0}{blue}, respectively.}
\label{tab:restoration_metrics}
\renewcommand{\arraystretch}{0.75}
\setlength{\tabcolsep}{15pt}
\newcommand{\txtbg}[2]{%
  \begingroup
  \setlength{\fboxsep}{1pt}%
  \colorbox[HTML]{#1}{\strut #2}%
  \endgroup
}
\resizebox{\textwidth}{!}{
\begin{tabular}{lcccccc}
\toprule
\textbf{Method} 
& \textbf{PSNR}$\uparrow$ 
& \textbf{SSIM}$\uparrow$ 
& \textbf{LPIPS}$\downarrow$ 
& \textbf{NIQE}$\downarrow$ 
& \textbf{PIQE}$\downarrow$ 
& \textbf{TOPIQ-NR}$\uparrow$ \\
\midrule

Bicubic 
& 21.76 
& 0.5323 
& 0.1721 
& 11.0904 
& 43.7747 
& 0.4173 \\

DiffVSR \cite{li2025diffvsr}
& 23.80 
& 0.7395 
& \txtbg{FFE4C1}{0.1011}
& 9.4150 
& 47.1095 
& 0.4651 \\

DUP \cite{chen2025blind}
& 23.67 
& 0.8096 
& 0.1665 
& 10.0552 
& 48.7218 
& 0.3836 \\

EDSR \cite{lim2017enhanced}
& 21.87 
& 0.7573 
& 0.2193 
& 17.3222 
& 57.1770 
& \txtbg{D4EFFB}{0.4848} \\

ZSSR-Cycle \cite{liu2021perception}
& 22.91 
& 0.7894 
& 0.2027 
& 17.5018 
& 50.1141 
& 0.4431 \\

\midrule

UltraDiff-DFUI
& \txtbg{FFE4C1}{25.27} 
& \txtbg{FFE4C1}{0.8229}
& 0.1843
& \txtbg{FFE4C1}{7.8461} 
& \txtbg{FFE4C1}{40.4669}
& 0.4624 \\

UltraDiff-ZEA
& \txtbg{FFBFBF}{25.62} 
& \txtbg{FFBFBF}{0.8484} 
& \txtbg{FFBFBF}{0.0909} 
& \txtbg{FFBFBF}{6.9604} 
& \txtbg{D4EFFB}{41.2375}
& \txtbg{FFBFBF}{0.5000} \\

UltraDiff-CelebA-HQ
& \txtbg{D4EFFB}{25.04}
& \txtbg{D4EFFB}{0.8209}
& \txtbg{D4EFFB}{0.1101}
& \txtbg{D4EFFB}{9.1370} 
& \txtbg{FFBFBF}{36.2833} 
& \txtbg{FFE4C1}{0.4920} \\

\bottomrule
\end{tabular}
}
\end{table*}

\subsection{Datasets}
For restoration evaluation, we use the EchoNet-LVH dataset~\cite{echolvh}, which contains high-resolution echocardiography videos with a resolution of 1024$\times$768 pixels and 90--200 frames per clip. Each video is cropped and resized to 256$\times$256, and the resulting frames are treated as reference HR videos. Since physically paired LR--HR ultrasound videos are rarely available, we follow the common full-reference evaluation protocol in ultrasound SR studies~\cite{Choi2018IUS,van2019deep,liu2021perception,chen2025blind}: LR observations are generated using 2$\times$ bicubic downsampling, followed by multiplicative Rayleigh speckle and additive Gaussian noise with a shared standard deviation varied from 0.02 to 0.10 to control the overall degradation strength. The same degraded LR inputs and HR references are used for all competing methods, ensuring a fair comparison under identical input conditions. Restoration quality is measured with the full-reference metrics PSNR, SSIM~\cite{wang2004image}, and LPIPS~\cite{zhang2018unreasonable}, and the no-reference metrics NIQE~\cite{mittal2012making}, PIQE~\cite{venkatanath2015blind}, and TOPIQ-NR~\cite{chen2024topiq}.

For downstream evaluation, we use the EchoNet-Dynamic dataset~\cite{echodynamic}, which contains real clinical echocardiography videos of 112$\times$112 pixels and 28--1002 frames per clip, together with reference EF labels. We select 100 videos using EF-stratified sampling so that the subset follows the EF distribution of the full dataset. This evaluation does not rely on synthetic degradations or HR references. Instead, it tests whether restoration improves functional measurement on real clinical videos, where the true degradation is unknown.

\subsection{Implementation Details}
\label{sec:implementation}
Our main UltraDiff configuration uses a CelebA-HQ pre-trained diffusion model as the frozen generative prior. This default is deliberate: the prior has never seen US data, so improvements can be attributed to the proposed formulation rather than to domain-specific training or any overlap between the training data, allowing us to test whether the proposed test-time optimization can adapt a generic image SR prior to ultrasound restoration. Unless otherwise noted, UltraDiff refers to this CelebA-HQ-prior configuration, which is used for ablation studies and downstream EF prediction.

To examine the sensitivity to the choice of frozen diffusion prior, we also instantiate the same UltraDiff optimization framework with two additional priors, denoted UltraDiff-DFUI and UltraDiff-ZEA, in the restoration comparison.

During test-time optimization, a DDIM sampler~\cite{song2020denoising} maps each latent seed to an HR estimate, while the diffusion model weights remain fixed and only the latent seed is optimized. The degradation operator $\mathcal{A}$ is implemented with bicubic downsampling, and the regularization weight is set to $\lambda=0.001$ in Eq.~\ref{eq:optim}. For the seed update strategy, the perturbation noise level is fixed at 0.1.

For the first frame of each video, we run up to 3000 optimization steps and enable early stopping after 2000 steps with patience 100. For subsequent frames, we reduce the maximum number of optimization steps to 2000 and enable early stopping after 800 steps with the same patience, since these frames are initialized from the propagated seed. For downstream evaluation, we use EchoCLIP~\cite{christensen2024vision} in a zero-shot setting to predict EF from SR videos. All input frames are resized to \(224 \times 224\) before inference. All experiments are implemented in PyTorch and conducted on a single NVIDIA GeForce RTX 3090 GPU with 24 GB memory. Baselines are evaluated on the same LR inputs using their public settings or official implementations.

\subsection{Comparison with Baselines}

We compare UltraDiff with five representative baselines from four groups: Bicubic interpolation, learning-based single-image SR (EDSR~\cite{lim2017enhanced}), video-based SR/restoration (DiffVSR~\cite{li2025diffvsr} and DUP~\cite{chen2025blind}), and ultrasound-specific SR (ZSSR-Cycle~\cite{liu2021perception}). This set covers interpolation, generic image SR, video restoration, and ultrasound-oriented SR. All methods use the same LR inputs and HR references, so differences reflect restoration behavior rather than degradation settings.

Fig.~\ref{fig:sr_visual} shows a qualitative comparison on a 2$\times$ SR example corrupted by mixed noise with std of 0.04. UltraDiff reconstructs clearer ventricular structures and produces lower errors in the zoomed regions than the baselines. For ultrasound restoration, improved sharpness should not come at the cost of distorted anatomical boundaries or unrealistic speckle texture.

The quantitative results in Table~\ref{tab:restoration_metrics} further support the visual comparison. The main UltraDiff-CelebA-HQ setting achieves 25.04 PSNR and 0.8209 SSIM, outperforming all non-UltraDiff baselines on both metrics. This result suggests that the proposed test-time diffusion-prior optimization can adapt a frozen image SR prior to echocardiography restoration, even when the prior is not trained specifically on ultrasound videos. The additional UltraDiff-DFUI and UltraDiff-ZEA variants use the same optimization framework with DFUI~\cite{dominguez2024diffusion} and ZEA~\cite{stevens2026zea} frozen priors, respectively, with ZEA achieving the strongest restoration scores. This shows that UltraDiff is not tied to a single checkpoint, while CelebA-HQ remains the default setting for ablation and downstream analyses.

\begin{table}[t]
\centering
\caption{Ablation study of UltraDiff components on restoration quality. Removing each component degrades performance.}
\label{tab:ablation}
\renewcommand{\arraystretch}{1.15}
\setlength{\tabcolsep}{5pt}
\begin{tabular}{lccccc}
\hline
\textbf{Variations} & \textbf{PSNR↑} & \textbf{SSIM↑} & \textbf{LPIPS↓} & \textbf{NIQE↓} & \textbf{PIQE↓} \\
\hline

MSE fidelity 
& 24.42 & 0.8087 & 0.1233 & 9.7698 & 38.0033 \\

w/o TV 
& 24.47 & 0.8076 & 0.1210 & 9.4159 & 38.6782 \\

w/o SUP 
& 24.13 & 0.7636 & 0.2934 & 11.2284 & 76.2607 \\

UltraDiff
& \textbf{25.04} & \textbf{0.8209} & \textbf{0.1101} & \textbf{9.1370} & \textbf{36.2833} \\

\hline
\end{tabular}
\end{table}

\begin{table}[t]
\centering
\caption{Downstream zero-shot EF prediction on the 100-video EchoNet-Dynamic subset after SR. MAE (EF \%), Acc@$\pm$5, and Recall@5$_{\pm5}$ are reported. UltraDiff denotes the default CelebA-HQ setting.}
\label{tab:echoclip_ef_transpose}
\footnotesize
\setlength{\tabcolsep}{4pt}
\begin{tabular}{lccc}
\toprule
\textbf{Method} & \textbf{MAE} $\downarrow$ & \textbf{Acc@$\pm$5} $\uparrow$ & \textbf{Recall@5$_{\pm5}$} $\uparrow$ \\
\midrule
Bicubic    & 11.00 & 28.00 & 67.00 \\
EDSR       & 10.60 & 26.00 & 58.00 \\
ZSSR-Cycle & 8.07  & 37.00 & 72.00 \\
DiffVSR    & 8.35  & 36.00 & 60.00 \\
DUP        & 7.31  & 44.00 & 72.00 \\
UltraDiff  & \textbf{6.37} & \textbf{53.00} & \textbf{74.00} \\
\bottomrule
\end{tabular}
\end{table}

\begin{figure}[t]
\centering
\includegraphics[width=0.7\linewidth]{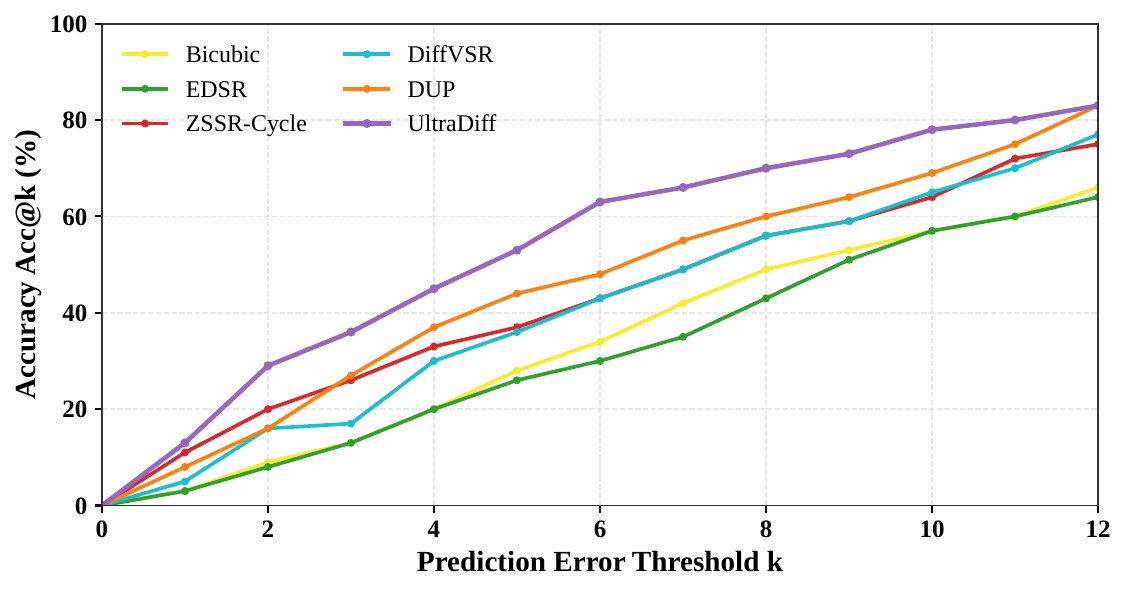}
\caption{Accuracy-error threshold curves for downstream EF prediction on the 100-video EchoNet-Dynamic subset. $\mathrm{Acc@}k$ measures the percentage of cases whose absolute EF error is within $k$ EF points. Higher curves indicate more reliable EF estimation across different error tolerances.}
\label{fig:ef_acc_curve}
\end{figure}

\begin{figure*}[!t]
\centering

\includegraphics[width=\linewidth]{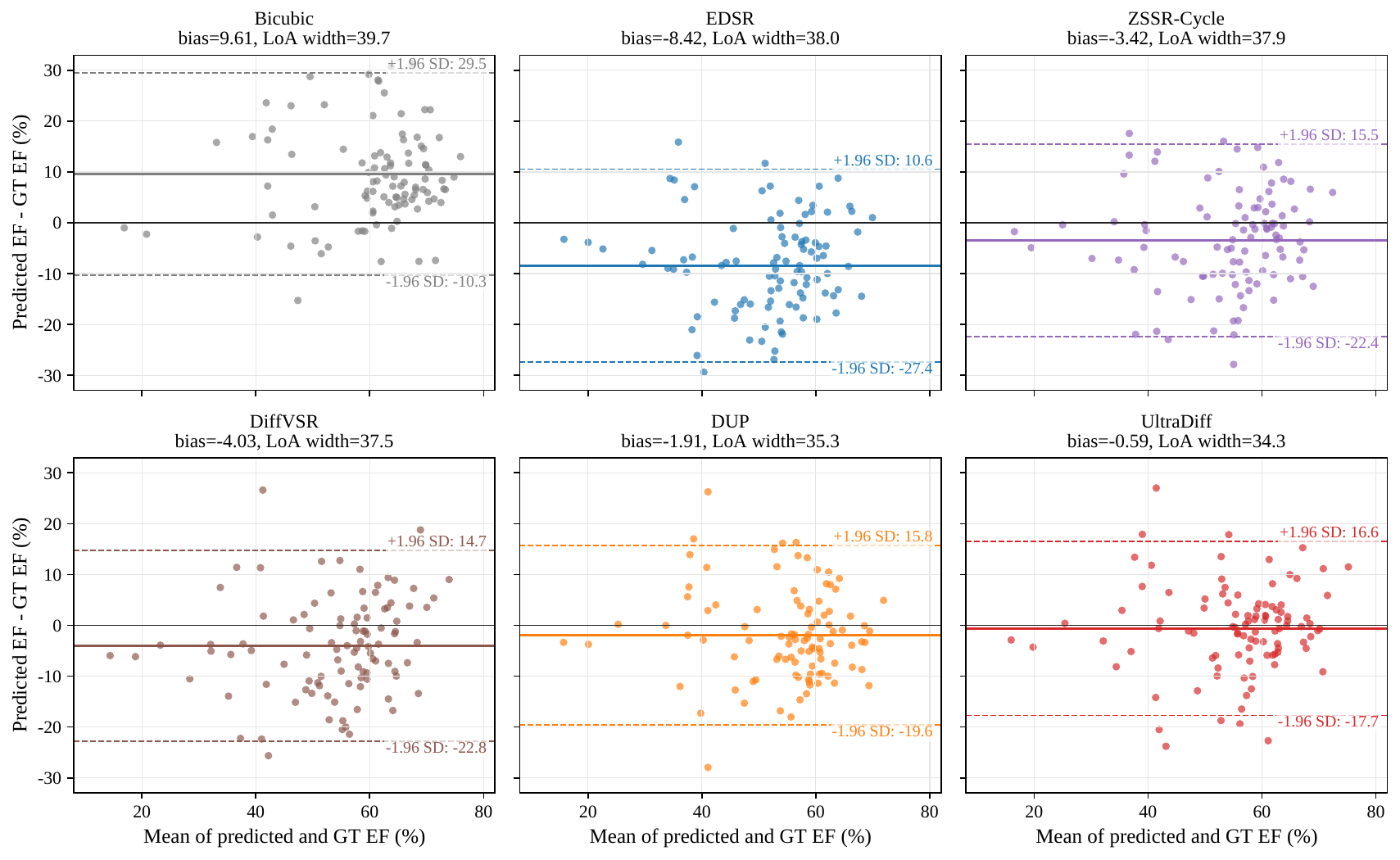}
\caption{Bland--Altman analysis of downstream EF prediction on the 100-video EchoNet-Dynamic subset. The solid line denotes the mean prediction bias, and the dashed lines denote the 95\% limits of agreement
($\mathrm{bias}\pm1.96\,\mathrm{SD}$). UltraDiff shows the smallest bias and the narrowest agreement range among the compared SR methods.}
\label{fig:ef_bland_altman}

\vspace{0.6em}

\includegraphics[width=\linewidth]{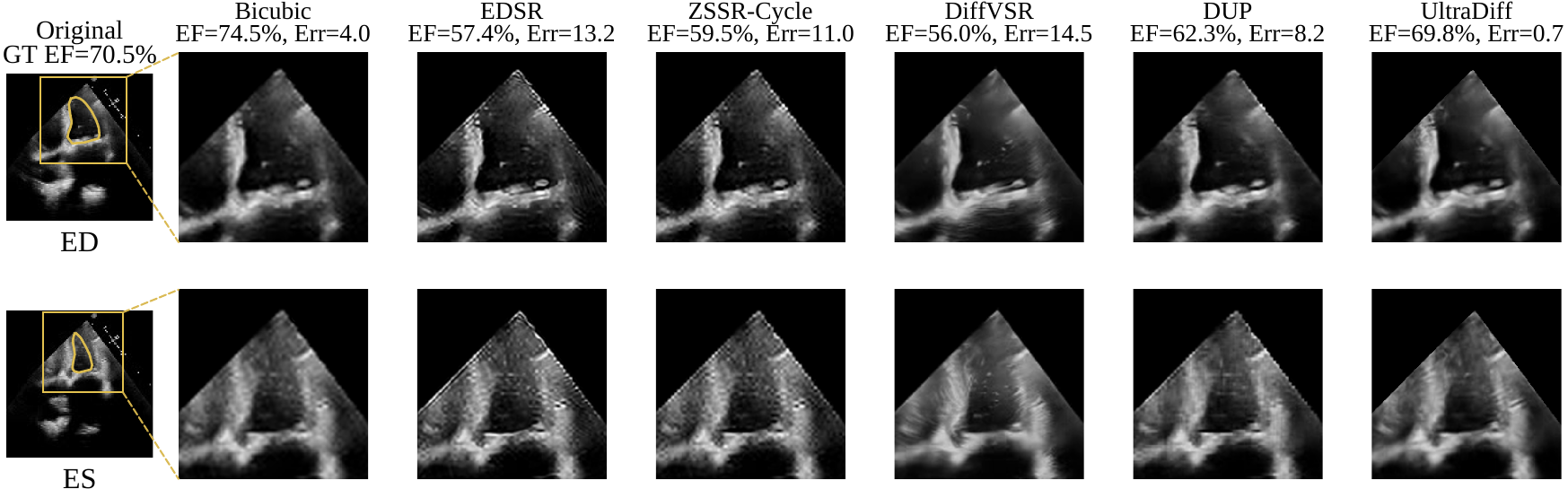}
\caption{Representative end-diastolic/end-systolic (ED/ES) visualization for downstream EF prediction. The original LR frames are shown with ground-truth left ventricular (LV) contours, and the dashed boxes indicate the LV regions enlarged for each 2$\times$ SR method. UltraDiff provides a clearer LV region while better preserving the ED/ES structural difference.}
\label{fig:ed_es_roi_callout}

\end{figure*}

\subsection{Robustness and Runtime Analysis}

To evaluate robustness under different degradation strengths, Fig.~\ref{fig:robust2noise} reports PSNR and SSIM as the mixed-noise level increases from 0.02 to 0.10. UltraDiff maintains higher reconstruction quality and a more concentrated distribution, suggesting that the speckle-aware fidelity term and diffusion prior stabilize restoration under stronger degradation.

UltraDiff is a test-time optimization method intended for offline restoration. Under the setup in Section~\ref{sec:implementation}, it restores a frame in about 83\,s on average, comparable to DUP (77\,s) and faster than DiffVSR (93\,s) and ZSSR-Cycle (152\,s), but slower than feed-forward baselines such as Bicubic and EDSR. The cost is dominated by per-frame latent optimization, while SUP warm-start and early stopping help reduce the cost for subsequent frames. This trade-off is suitable for retrospective EF analysis, where offline processing is acceptable.

\subsection{Ablation Studies}

We conduct ablation studies to isolate the contribution of each component in UltraDiff. Specifically, we replace the proposed fidelity term with an MSE-based fidelity, remove TV regularization, and disable the seed update strategy. As shown in Table~\ref{tab:ablation}, each modification degrades performance, confirming that the fidelity term, regularization, and temporal seed propagation all contribute to restoration quality.

Replacing $\ell_s$ with the MSE reduces PSNR from 25.04 to 24.42\,dB and increases LPIPS from 0.1101 to 0.1233, consistent with Propositions~\ref{prop:fidelity} and~\ref{prop:stab}: the homoscedastic MSE over-weights bright, high-variance regions, whereas $\ell_s$ measures residuals relative to the local signal level. Removing TV yields a comparable PSNR drop (24.47\,dB) and a higher PIQE, reflecting its role in promoting piecewise-constant structures: it suppresses residual fluctuations within homogeneous regions, such as the blood pool and myocardium, while preserving tissue boundaries. Disabling SUP causes the largest degradation, especially in SSIM, LPIPS, and PIQE: without the warm start, each frame starts from an unrelated seed whose initial estimate is far from the target, so the optimization is less likely to reach a good solution within the budget, in line with the bounded-discrepancy argument in Section~\ref{sec:sup}.

\subsection{Downstream Evaluations}
\label{downstream}
Restoration metrics alone do not fully indicate whether an enhanced ultrasound video is useful for clinical analysis. We therefore evaluate all SR methods on the 100-video EchoNet-Dynamic subset for zero-shot ejection fraction (EF) prediction using EchoCLIP~\cite{christensen2024vision}. As reported in Table~\ref{tab:echoclip_ef_transpose}, we use mean absolute error (MAE) in EF percentage points, Acc@$\pm$5, and Recall@5$_{\pm5}$. Acc@$\pm$5 measures the percentage of predictions within $\pm$5 EF points, while Recall@5$_{\pm5}$ measures whether one of the top-5 EF candidates falls within the same tolerance.

UltraDiff achieves the best performance across all three metrics, with an MAE of $6.37$, Acc@$\pm$5 of $53.00\%$, and Recall@5$_{\pm5}$ of $74.00\%$. Compared with Bicubic interpolation, UltraDiff reduces MAE from $11.00$ to $6.37$ and improves Acc@$\pm$5 from $28.00\%$ to $53.00\%$. It also reduces MAE by $12.9\%$ over DUP, the strongest non-UltraDiff baseline. Fig.~\ref{fig:ef_acc_curve} further shows that UltraDiff achieves the highest or tied-highest $\mathrm{Acc@}k$ across error tolerances, indicating that the improvement is not limited to a single threshold. The Bland--Altman analysis in Fig.~\ref{fig:ef_bland_altman} shows the smallest mean bias ($-0.59$ EF points) and the narrowest 95\% limits of agreement, indicating closer agreement between predicted and reference EF values in this evaluation.

Finally, Fig.~\ref{fig:ed_es_roi_callout} provides a representative end-diastolic/end-systolic (ED/ES) case. The original LR frames are overlaid with ground-truth left ventricular (LV) contours, and the corresponding LV regions are enlarged across methods. Since EF estimation depends on the structural change between ED and ES frames, a useful restoration method should enhance the LV region without distorting this temporal anatomical difference. UltraDiff provides a clearer LV region while better preserving the ED/ES structural difference, consistent with its improved downstream EF prediction performance.

\section{Discussion}
\label{sec:discussion}

\textit{Role of the priors.} The DM enters~\eqref{eq:map} not as an explicit density but through the range constraint $I\in\mathcal{R}(\mathcal{M})$ and the inductive bias of the pre-trained denoiser. TV is an explicit regularizer that favors piecewise-constant structures: it suppresses residual noise within homogeneous regions, such as the blood pool and myocardium, while preserving sharp tissue boundaries, and it stabilizes the ill-posed SR problem. The two priors are complementary: TV regularizes homogeneous regions, whereas fine tissue texture is supplied by the DM.

\textit{Choice of prior.} Because $K$ is small and $z$ is optimized without constraint, $\mathcal{R}(\mathcal{M})$ is in practice not restricted to the images on which the DM was trained (e.g., faces for CelebA-HQ). Although natural and medical images differ greatly at the whole-image scale, the statistics of local features, such as heavy-tailed gradient distributions, are similar across domains, and diffusion-based reconstruction outside the training distribution is driven primarily by such local features~\cite{glaszner2026bigger}. Accordingly, DMs trained on natural images, including CelebA-HQ, outperform medical-image-trained DMs in out-of-distribution MRI and CT reconstruction~\cite{glaszner2026bigger} and enable training-free enhancement across medical modalities, including ultrasound~\cite{fei2025diffusion}. This reliance on local regularity also mitigates hallucination of domain-specific content: global structures of the training domain are not imposed, because the data term ties $\mathcal{A}\mathcal{M}(z)$ to the observation at every pixel while the prior mainly contributes local regularity. The natural-image default thus shows that the gains stem from the formulation rather than domain-specific training.

\textit{Extensions of the measurement model.} UltraDiff operates on displayed intensities, which makes it directly applicable to standard clinical archives such as EchoNet without access to raw data. The log-residual form of the fidelity (Proposition~\ref{prop:fidelity}(i)) is naturally compatible with the logarithmic display of US images, and the scale-family formulation accommodates other speckle statistics, e.g., for partially developed speckle~\cite{sivaanpu2025speckle}, simply by changing $f$. The EF results on real videos, whose degradation is unknown, indicate robustness beyond the operator. Promising extensions include envelope or radio-frequency data where available, spatially correlated speckle models, and joint estimation of $\mathcal{A}$ for fully blind restoration.

\textit{Future directions.} The modular design of UltraDiff allows each component to be upgraded independently. On the prior side, video DMs with built-in temporal modeling and ultrasound foundation DMs can be plugged in without retraining. On the regularization side, higher-order terms~\cite{chan2000high} and total generalized variation~\cite{bredies2010total}, which favor piecewise-smooth rather than piecewise-constant solutions, can further refine smoothly varying tissue regions. The per-frame optimization can be accelerated through diffusion distillation, consistency models, fewer sampling steps, or amortized seed prediction toward near-real-time operation. Building on the EF results reported here, future validation will cover higher scale factors, other anatomies such as obstetric and musculoskeletal US, dedicated temporal-consistency metrics, and clinical reader studies.

\section{Conclusion}
We presented UltraDiff, a self-supervised US video restoration framework that uses a frozen pre-trained diffusion model as a plug-in prior. By optimizing latent seeds under a range-constrained MAP-inspired formulation with a speckle-aware, variance-adaptive fidelity and a temporal seed update, UltraDiff performs joint super-resolution and despeckling without paired HR data or model retraining. Experiments on two public echocardiography datasets demonstrate improved restoration quality, robustness to mixed noise, and better zero-shot EF prediction than representative SR baselines, supporting diffusion-prior-based self-supervised restoration for retrospective US video analysis.

\section*{Acknowledgments}
This work was supported in part by the Start-up Grant of City University of Hong Kong (Dongguan).

\bibliographystyle{unsrt}  
\bibliography{reference}

\end{document}